\documentclass{article}

\usepackage{enumitem}
\usepackage{graphicx}

\usepackage{algpseudocode}
\usepackage{amsmath}

\usepackage{algorithm}

\PassOptionsToPackage{numbers}{natbib}

\usepackage[final]{neurips_2023} 

\usepackage[utf8]{inputenc} 
\usepackage[T1]{fontenc}    
\usepackage{hyperref}       
\usepackage{url}            
\usepackage{booktabs}       
\usepackage{amsfonts}       
\usepackage{nicefrac}       
\usepackage{microtype}      
\usepackage{xcolor}         

\definecolor{customRed}{RGB}{235,0,0}
\definecolor{customGreen}{RGB}{00,156,0}
\definecolor{customBlue}{RGB}{31, 136, 228}
\definecolor{customPink}{RGB}{216, 27, 96}

\title{Automated clinical coding using off-the-shelf\\  large language models}

\author{%
  \textbf{Joseph S. Boyle}\textsuperscript{1,2},
  \textbf{Antanas Kascenas}\textsuperscript{1,3},
  \textbf{Pat Lok}\textsuperscript{1,4},
  \textbf{Maria Liakata}\textsuperscript{2,5,6},
  \textbf{Alison Q. O'Neil}\textsuperscript{1,3} \\
  \\
  \textsuperscript{1}Canon Medical Research Europe,
  \textsuperscript{2}Queen Mary University of London,\\
  \textsuperscript{3}University of Edinburgh,
  \textsuperscript{4}Anglia Ruskin University, \\
  \textsuperscript{5}The Alan Turing Institute,
  \textsuperscript{6}University of Warwick \\
  \\
  \texttt{joseph.boyle@mre.medical.canon} \\
}

\begin{document}

\maketitle

\begin{abstract}
The task of assigning diagnostic ICD codes to patient hospital admissions is typically performed by expert human coders.
Efforts towards automated ICD coding are dominated by supervised deep learning models. However, difficulties in learning to predict the large number of rare codes remain a barrier to adoption in clinical practice. 
In this work, we leverage off-the-shelf pre-trained generative large language models (LLMs) to develop a practical solution that is suitable for zero-shot and few-shot code assignment, with no need for further task-specific training.
Unsupervised pre-training alone does not guarantee precise knowledge of the ICD ontology and specialist clinical coding task, therefore we frame the task as information extraction, providing a description of each coded concept and asking the model to retrieve related mentions.
For efficiency, rather than iterating over all codes, we leverage the hierarchical nature of the ICD ontology to sparsely search for relevant codes.
We validate our method using Llama-2, GPT-3.5 and GPT-4 on the CodiEsp dataset of ICD-coded clinical case documents. Our tree-search method achieves state-of-the-art performance on rarer classes, achieving the best macro-F1 of 0.225, whilst achieving slightly lower micro-F1 of 0.157, compared to 0.216 and 0.219 respectively from PLM-ICD.
To the best of our knowledge, this is the first method for automated clinical coding requiring no task-specific learning.
\end{abstract}

\section{Introduction}

International Classification of Disease (ICD) coding is the task of assigning ICD codes to episodes of patient care e.g. hospital stays. Assigned codes may be used for billing, audit, resource management, epidemiological study, measurement of treatment effectiveness, and other purposes. Coding is usually a manual process, performed by specialists via inspection of a patient's medical documentation. This is time-consuming and error-prone; with one study estimating the costs related to medical coding to be billions of dollars per year in the US alone \citep{kaur_ai-based_2023}. Automation of clinical coding has been pursued since the late 1990s \citep{de_lima_hierarchical_1998}, with deep learning techniques currently dominating \citep{edin_automated_2023}.

A challenge for automation is the large number of codes that must be predicted. The ICD-10-CM ontology contains 96,000 distinct codes, of which 73,000 are assignable \citep{world_health_organization_icd-10_2004}. The remaining 23,000 codes are higher-level parent codes corresponding to broader categories of codes. Many assignable codes are rare, and there may be few or no examples in any given training set. Moreover, there is considerable conceptual overlap between ICD codes. This poses difficulties for supervised learning, which relies on large volumes of labelled training data to learn distinct representations for closely related concepts. Compared to other text classification tasks, ICD coding is a challenging problem; quantitatively, the state-of-the-art achieves a micro-F1 of 0.585 and macro-F1 of 0.211 on the MIMIC-IV ICD coding benchmark \citep{edin_automated_2023}. 

In this paper, we explore an alternative approach using off-the-shelf generative LLMs which have been trained using self-supervised learning on trillions of tokens \citep{brown_language_2020, touvron_llama_2023}. Recent models have proven powerful for medical tasks such as question answering, summarisation and clinical information retrieval \citep{agrawal_large_2022}, demonstrating good performance on clinical text and notably achieving an 86.5\% (pass) score on the MedQA dataset of questions from the US Medical License Examination \citep{singhal_towards_2023}. We consider how to leverage generative LLMs to perform the task of ICD coding even in the absence of labelled training examples i.e. with no task-specific training. Leveraging the hierarchical nature of the ICD ontology, we propose a novel LLM-based solution that formulates the task as a dynamic sequence of searches for clinical entities, with the sequence corresponding to paths followed through the tree to all relevant assignable ``leaf'' codes. In this framework, the LLM is prompted to assess the relevance of each branch of the tree based on its text description. Our contributions are as follows:
\begin{enumerate}
    \item We create the first method for ICD coding requiring no task-specific training or fine-tuning.
    \item We demonstrate that LLMs have some out-of-the-box ICD coding abilities.
    \item We propose a method that avoids reliance on model knowledge of the target coding ontology via the injection of information into the LLM prompt and the application of a novel search strategy conceptually similar to a multi-label decision tree. We show empirically that this tree-search strategy improves model performance on rare codes.
\end{enumerate}

\section{Related Work}

Models building on pre-trained encoder transformers such as BERT currently lead performance benchmarks for ICD coding \citep{edin_automated_2023, huang_plm-icd_2022}. However, prediction for codes with few or no training examples remains difficult. Song et al. proposed a novel GAN-based feature generation approach for zero-shot labeling by exploiting the ontological structure of ICD to extrapolate features learned from seen sibling and parent codes \citep{song_generalized_2020}. Lu et al. proposed aggregating multiple label graphs (the original code ontology, the semantic similarity graph computed from their descriptions, and a label co-occurrence frequency matrix) as a per-label pre-processing step \citep{lu_multi-label_2020}. Recently, Yang et al. conducted a study into few-shot ICD coding by reformulating the classification problem as masked language modelling one (a `cloze task') such that the model learns to replace the masked tokens in the input prompt with binary predictions \citep{yang_knowledge_2022}. Each of these approaches require labelled data for model training. In contrast, we seek to develop zero-shot learning methods for this task.

\section{Data}

The open access MIMIC dataset is the most commonly used benchmark for this task \citep{kaur_ai-based_2023}, however the conditions of use prohibit sharing of data with third parties (such as OpenAI). Therefore, we chose to use \textbf{CodiEsp}, a publicly available dataset which formed the basis of the \textit{eHealth CLEF 2020 Multilingual Information Extraction Shared Task}, a competition for automated clinical coding \citep{miranda-escalada_overview_2020}. In this competition, 1000 span-level expert-annotated case notes were released in Spanish, alongside machine-translated English versions. We evaluate on the competition test set but abstract the span-level labels into document-level labels as would be available in real-world clinically coded data. This comprises 250 case note documents from 250 unique patients, covering 1767 distinct ICD-10 codes (2.4\% of the ICD-10-CM codeset).

Inspection of the translated documents revealed errors such as failure to translate Spanish terms for drugs. Since translation errors could hamper diagnostic coding performance, we re-translated the documents using GPT-3.5, which reduced errors and yielded modest performance improvements for all models during experimentation. We henceforth refer to this as the `CodiEsp-English' dataset. Lastly, we remove the small number of ground truth labels not assignable in the ICD-10-CM.

\section{Method}

We describe our methods and their implementation below.

\begin{figure*}[!thbp]
  \centering
  \includegraphics[width=\textwidth]{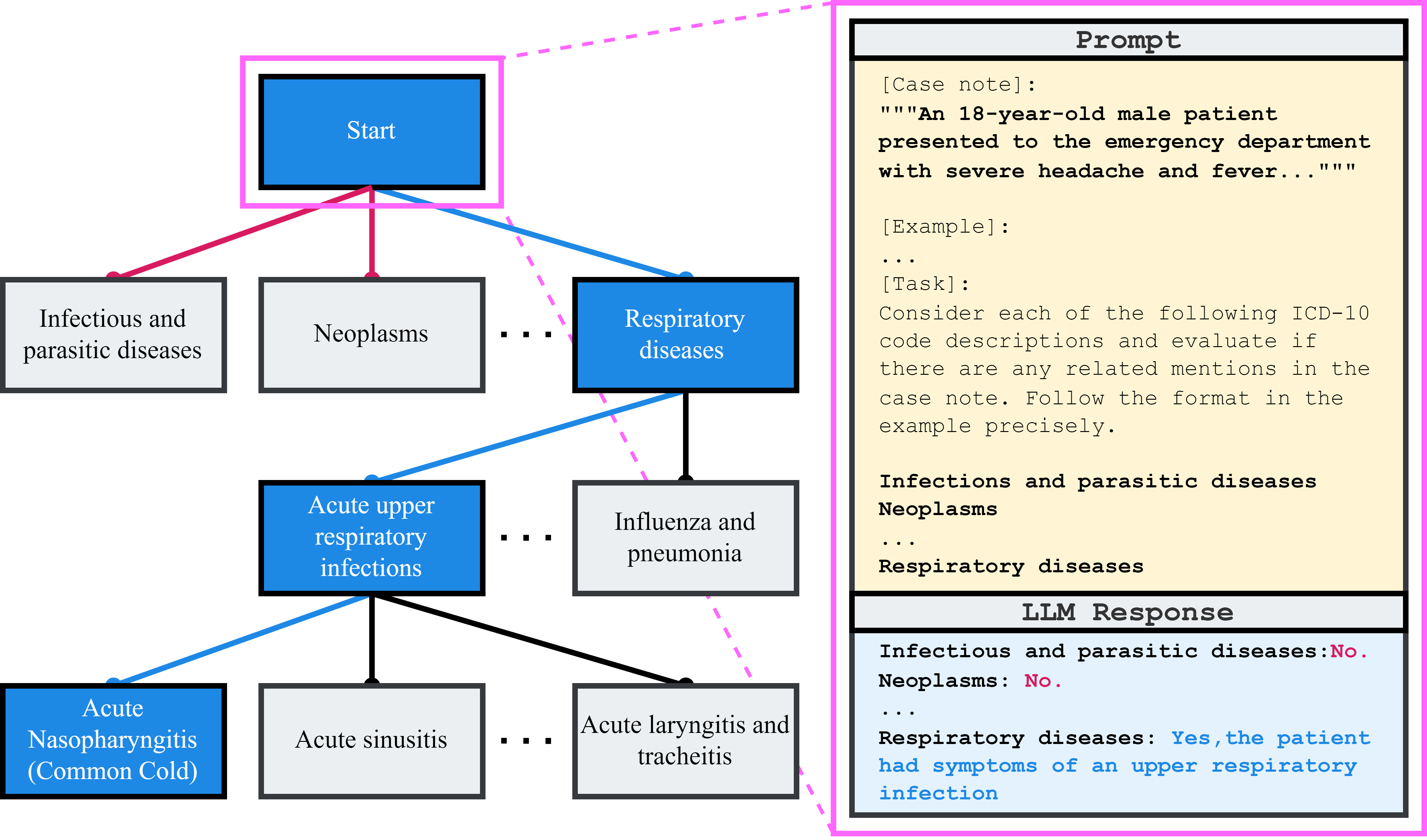}
  \caption{Illustration of our LLM-guided tree-search method, as applied to assign an example code \textbf{\textcolor{customBlue}{``Acute nasopharyngitis (Common Cold)''}} to CodiEsp document \texttt{S0212-71992006000100006-1}, using GPT-4. Left: ICD-10 diagnostic tree, showing the codes traversed to reach the correct assignable ``leaf'' code. Right: Abridged prompt containing the case note; we colour the LLM model's response with \textbf{\textcolor{customBlue}{blue}} and \textbf{\textcolor{customPink}{pink}} indicating the positively and negatively predicted codes, respectively.}
  \label{fig:icd-tree-search}
\end{figure*}

\paragraph{Information retrieval task framing:} We hypothesised that LLMs might be capable ICD coders `out-of-the-box'. Whilst the prompt ``\texttt{You are a clinical coder, consider the case note and assign the appropriate ICD codes}'' elicits the correct behaviour from GPT-4, the responses contained inaccuracies such as mismatched codes and descriptions (see full prompt in Appendix \ref{app:clinical-coder-prompt}). For instance, the model assigned the code: ``C63.2 - Malignant neoplasm of \textbf{\textcolor{customPink}{left testis}}' where the true description for the code ``C63.2'' is `Malignant neoplasm of \textbf{\textcolor{customBlue}{scrotum}}''.

We therefore reframe the problem as information retrieval and prompt the LLM to retrieve `mentions' of candidate codes from the case note. Taking inspiration from Yang et al. \citep{yang_knowledge_2022}, we include the descriptions in the prompt, as shown in Figure \ref{fig:icd-tree-search}. 
If a relevant mention is found, the code is predicted.

\paragraph{LLM guided tree-search:} Determining which ICD code descriptions to employ in the information retrieval prompt is non-trivial as the ICD includes thousands of codes. The ICD diagnostic ontology is a tree, with `is a' (hyponymic) semantics relating child and parent codes. For instance, the child code \emph{Acute Nasopharyngitis} is an \emph{Upper Respiratory Infection} (parent code). We exploit the tree-like structure of the ontology to efficiently search for relevant ICD codes, using the LLM to decide at each decision point which paths to explore. Starting at the root of the tree, the model selects relevant chapters to explore, as shown in the prompt-response pair in Figure \ref{fig:icd-tree-search}. This is performed recursively until the set of candidate paths to explore is exhausted, at which point the set of relevant labels (leaf codes) is returned. For instance the search path shown in Figure \ref{fig:icd-tree-search} would add the label `Acute Nasopharyngitis' to the set of predicted labels. The exact algorithm is described in Appendix \ref{app:tree-search-algorithm}.

For a relational model of the ICD diagnostic tree, we used the open-source \texttt{simple\_icd\_10\_cm} library\footnote{\url{https://github.com/StefanoTrv/simple_icd_10_CM}}. The LLMs used to guide the tree-search are Llama-2 (70B-chat), GPT-3.5, and GPT-4 \citep{brown_language_2020, touvron_llama_2023}. Llama-2 was accessed via the hosted DeepInfra service \footnote{\url{https://deepinfra.com/}}, and GPT-3.5/4 via OpenAI's API. The \texttt{06-13} revisions of both GPT-3.5 and GPT-4 were used. To get the most deterministic model outputs possible, the \texttt{temperature} parameter was set to it's minimum value; $0$ and $0.001$ for GPT and Llama respectively (smaller values did not work for Llama).

\paragraph{Output parsing:} Resolving the LLM generated text into per-code predictions is performed by processing the text as a set of lines. These lines are greedily matched, starting with the longest code description in the current prompt. This is relevant as in some instances there are code descriptions which are substrings of another code descriptions. For instance, the description of A48.1 is `Legionnaires' disease', a substring of A48.2, `Nonpneumonic Legionnaires' disease'. To avoid erroneously matching model outputs we first string match for a line corresponding to the longer description.

\paragraph{Baselines:} We report baseline results from the Pretrained-Language-Model framework (PLM-ICD) \citep{huang_plm-icd_2022} which is the state-of-the-art (SOTA) model for the task of ICD coding \citep{edin_automated_2023}. It combines BERT as a text encoder model with per-label attention and per-label binary classification heads.
Due to the limited size of the CodiEsp training dataset (500 case notes), we use the PLM-ICD provided model weights learnt on the MIMIC-IV dataset \citep{edin_automated_2023} \footnote{
    \url{https://github.com/JoakimEdin/medical-coding-reproducibility}
}. This represents a realistic transfer learning scenario.

Our second `clinical coder' baseline is the previously described out-of-the-box approach of prompting the model to act as a clinical coder, without providing further information about ICD.
We identify the returned ICD codes in two ways: a) by matching the alpha-numeric codes themselves and b) by matching their natural language descriptions.

\section{Results}

\paragraph{Main Results:} Performance metrics are shown in Table \ref{tab:codiesp-results}. Our evaluation metrics are micro (instances weighted equally) and macro (classes weighted equally) precision, recall and F1 scores. We consider macro-F1 to be the most representative metric for performance on rare / zero-shot codes, as it weights all classes equally.
The PLM-ICD baseline performed well, achieving the best micro-F1 score. Our out-of-the-box single prompt `You are a clinical coder...' method demonstrated similar performance on micro-metrics but poorer performance on macro-metrics compared to our proposed tree-search method, which achieved the highest macro-F1 when GPT-4 was used.






\begin{table*}[!th]
\centering
    \begin{tabular}{lllllllll}
        \toprule
        &                                           \multicolumn{3}{c}{Micro}          &  \multicolumn{3}{c}{Macro} \\
        \toprule
        \textbf{Model}           & \textbf{Rec.} & \textbf{Prec.} & \textbf{F1} & \textbf{Rec.} & \textbf{Prec.} & \textbf{F1} \\
        \midrule
        PLM-ICD                 & 0.213            & 0.225            & \textbf{0.219}                     & 0.244            & 0.237        & 0.216 \\
        \midrule
        \multicolumn{7}{l}{Clinical coder (match codes):} \\
        Llama-2                 & 0.011             & 0.033            & 0.016                    & 0.007          & 0.011              & 0.006  \\
        GPT-3.5                 & 0.163             & 0.155            & 0.159                    & 0.149          & 0.161              & 0.136\\
        GPT-4                   & 0.242             & 0.161            & 0.193                    & 0.219          & 0.214              & 0.195 \\
        \midrule
        \multicolumn{7}{l}{Clinical coder (match descriptions):} \\
        Llama-2                 & 0.037             &   \textbf{0.282} & 0.065                    & 0.034           & 0.061            & 0.040 \\
        GPT-3.5                 & 0.147             & 0.168            & 0.157                    & 0.135           & 0.155            & 0.128 \\
        GPT-4                   & 0.217             & 0.166            & 0.188                    & 0.187           & 0.190             & 0.169 \\
        \midrule
        \multicolumn{7}{l}{Tree-search:} \\
        Llama-2                 & 0.173             & 0.039             & 0.064                    & 0.197          & 0.152              & 0.144 \\
        GPT-3.5                 & 0.206             & 0.159             & 0.179                    & 0.220          & \textbf{0.241}     & 0.208 \\
        GPT-4                   & \textbf{0.331}    & 0.087             & 0.138                    & \textbf{0.381} & 0.190              & \textbf{0.225} \\
        \bottomrule
    \end{tabular}
      \caption{Results for the supervised PLM-ICD model in a transfer-learning context;
    the single prompt `you are a clinical coder', and the LLM tree-search algorithm on the CodiEsp English dataset.}
    \label{tab:codiesp-results}
\end{table*}

\paragraph{Level-wise Analysis} In Table \ref{tab:level-analysis} we show the performance of our GPT-4 tree-search method at each level of the ICD ontology. Numbers are cumulative i.e. performance at a lower level of the tree includes prediction failures at higher levels of the tree, allowing us to observe the level-wise degradation of performance. However, since assignable leaf codes (class labels) can occur from the `category' level onwards, meaning some codes are ``dropped'' before the extension levels, it is not possible to directly relate the results from results tables \ref{tab:codiesp-results} and \ref{tab:level-analysis}. It can be seen that both micro- and macro-recall drop to approximately 50\% already by the second `Block' level, with approximately 40\% reaching the assignable levels (Subcategory, Extension I, extension II).


\begin{table*}[!th]
    \centering
    \begin{tabular}{lrrrrrr}
        \toprule
        &                                           \multicolumn{3}{c}{Micro}          &  \multicolumn{3}{c}{Macro}             \\
                        & Rec.                 &   Prec.   & F1        & Rec.    &   Prec.   & F1                        \\
        \toprule
            Chapter      &         0.835 &            0.677 &           0.748 &          0.778 &            0.668 &           0.693 \\
            Block        &         0.501 &            0.321 &           0.392 &          0.529 &            0.452 &           0.431 \\
            Category     &         0.401 &            0.201 &           0.268 &          0.448 &            0.392 &           0.364 \\
            Subcategory  &         0.343 &            0.110 &           0.167 &          0.418 &            0.370 &           0.348 \\
            Extension I  &         0.347 &            0.064 &           0.108 &          0.362 &            0.319 &           0.306 \\
            Extension II &         0.192 &            0.028 &           0.049 &          0.216 &            0.212 &           0.199 \\
        \bottomrule
    \end{tabular}
    \caption{Cumulative performance metrics, computed on the GPT-4 tree-search CodiEsp English test set predictions.}
    \label{tab:level-analysis}
\end{table*}

\paragraph{Prediction of Mutually Exclusive Codes} For the tree-search strategy, a common failure mode was predicting several similar leaf codes, leading to poor precision. For instance, GPT-4 tree-search predicted codes `B27.89' and `B27.80' for document \texttt{S0212-71992006000100006-1}: that is, mononucleosis with \textit{and} without complications, when these labels are clearly mutually exclusive.

\section{Discussion}

Reported performance in the original CodiEsp competition was comparatively high, with a dictionary-based method winning the competition with a micro-F1 score of 0.687. However, in real clinical data there are no span-level labels to learn from. Indeed, as we have seen for many classes there are not even samples to learn from, hence our choice of a zero-shot approach. Whilst the CodiEsp models may perform well on the small fraction of labels represented in the dataset, they are not generalisable to the remaining labels, whereas our tree-search method is truly general in this sense.

The best strategy for prompting models is not well understood. We could not find a common prompt that both Llama-2 and GPT-3.5/4 would adhere to, which forced us to employ separate prompt templates for each model. Due to its relative newness, we have less experience developing prompts for Llama-2 than for GPT models, which may somewhat account for the poor performance of Llama-2 in our experiments.


The advantage of our method (sparse exploration of the ICD tree) is also its potential weakness, in the sense that false negative errors can compound at each level. For instance, for a `subcategory' leaf code to be predicted, each of its parent, grandparent and great-grandparent codes must also be predicted; if any one of those is not predicted then the code is unreachable. This is illustrated in Table \ref{tab:level-analysis} which shows that ancestors of many true codes are not correctly predicted, thus the codes cannot be predicted, even if the model deems them ``relevant'' when directly provided with their descriptions.

There are a few promising avenues for future research. Our current approach has the advantage of being generic; we use the exact same prompt for every node and can operate with any tree of concepts for which each concept has an associated text description. However, to achieve better accuracy for ICD coding, we might introduce consideration of the ICD coding rules to allow/disallow legitimate code assignment combinations (i.e. mutually exclusive codes), or try to improve level-wise accuracy by tailoring the prompt for different tree levels (e.g. use a different prompt template for ``chapter'' prompts compared to ``subcategory'' prompts etc.).

\section{Conclusion}
In this paper we have presented a promising approach for performing ICD coding with generative LLMs without the need for task-specific training. 
We note that our first LLM method, the out-of-the-box `You are a clinical coder...' prompt, is heavily contingent on the model's pre-trained knowledge of the ICD-10 ontology and the specialist clinical coding task. Our tree-search method does not have this requirement, and thus could handle little/never seen rare codes, leading to better macro-averaged performance. In practice, this would allow handling of new codes such as the \texttt{UO7.1 - COVID-19} code introduced in February 2020, or whole ontology revisions  such as ICD-11 (introduced in January 2022 \citep{pezzella_icd11_2022}). We conclude that our LLM tree-search method and generative LLMs more broadly show exciting potential for zero-shot clinical coding. 

\medskip

{
\small

\newpage

\bibliography{references}
}

\appendix

\section{Appendix}



\vspace{20pt}
\subsection{Clinical coder prompt}
\label{app:clinical-coder-prompt}

Below we show the prompt used for the out-of-the-box `You are a clinical coder...' method.

\begin{figure}[h]
    \centering
    \includegraphics[width=\textwidth]{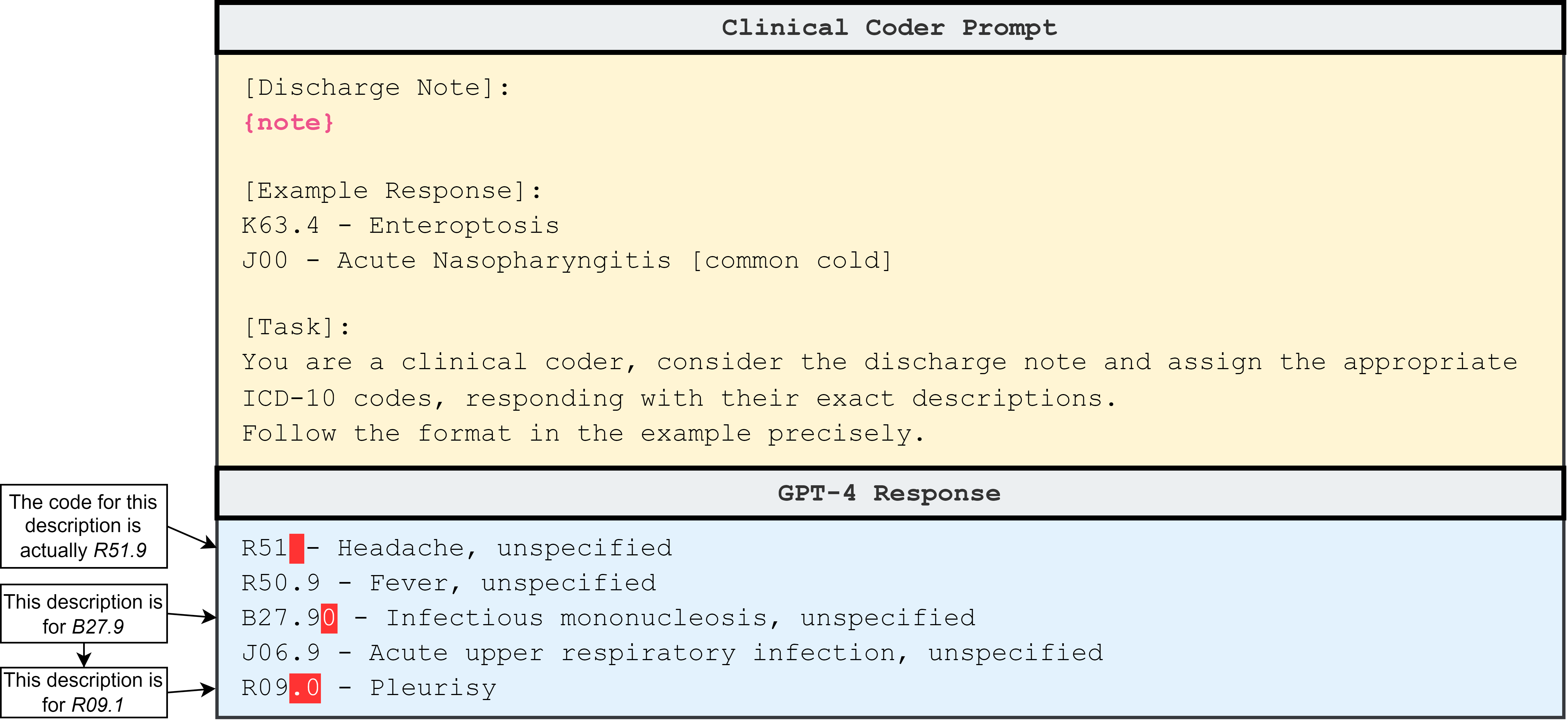}
    \caption{The `You are a clinical coder...' prompt used for both GPT and Llama models and a sample GPT-4 response for document \texttt{S1130-05582008000500007-1}. The discrepancies between the true ICD codes and code descriptions generated by GPT-4 are highlighted in red.}
    \label{fig:clinical-coder-prompt}
\end{figure}



\newpage
\subsection{LLM guided Tree-Search prompts}
\label{app:tree-search-prompts}

Below we show the two prompts used in our tree-search method for the GPT and Llama models respectively.

\begin{figure}[h]
    \centering
    \includegraphics[width=\textwidth]{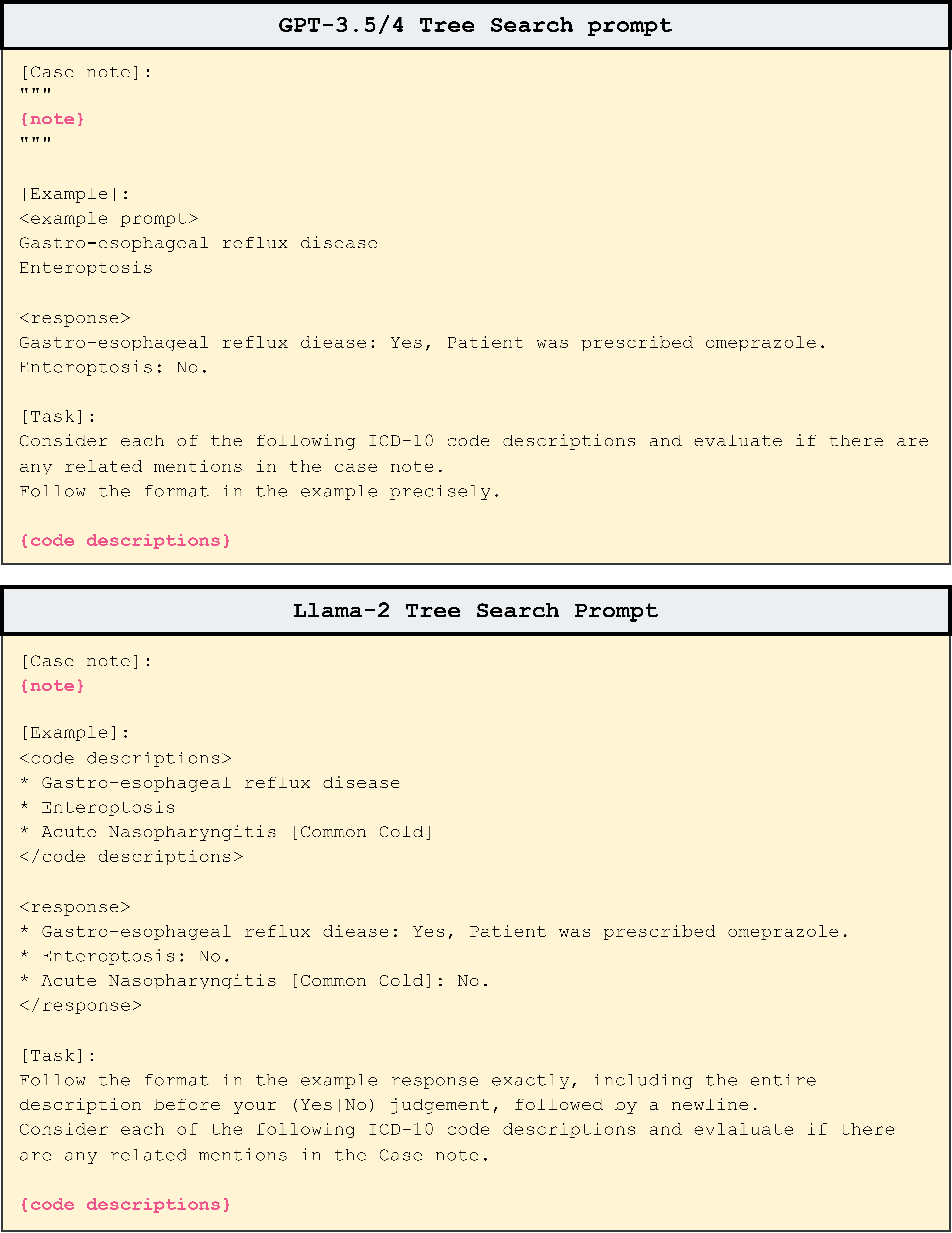}
    \caption{Tree-search prompt templates for GPT and Llama models.}
    \label{fig:tree-prompts}
\end{figure}

We found that Llama-2 exhibited poor adherence to the desired output format when prompts
developed for GPT were used. Differences between the prompts reflect experimental improvements in achieving model adherence to this output format. For example, Llama-2
preferred to respond with bullet points, so we adapted our prompt to encourage this, as it improved the parse-ability of the model response.


\subsection{Tree-search algorithm}
\label{app:tree-search-algorithm}

Below we describe the steps of the tree-search algorithm.

\begin{algorithm}[h]

\caption{Assign a set of ICD codes for a case note using LLM-guided tree-search.}
    \textbf{Requisites:}
    \begin{itemize}
        \item \textbf{tree}: an ICD ontology, with methods for accessing the natural language code descriptions,
            selecting the children of a code and checking if a given code is assignable (i.e a leaf code).
        \item \textbf{prompt\_template}: An LLM prompt which embeds the string variables `case\_note' and `code\_descriptions'.
        \item \textbf{llm\_api\_request}: a function accepting a case note and a set of code descriptions to be inserted into
            the prompt. Returns a text completion from the LLM.
        \item \textbf{match\_code\_descriptions}: a function which takes an LLM text completion and the set of candidate code descriptions and returns the codes which the model predicted as being relevant.
    \end{itemize}
    \begin{algorithmic}[1]
        \Function{search\_tree}{case\_note}
          \State assigned\_codes $\leftarrow$ ()
          \Comment{Predicted leaf codes}
          \State candidate\_codes $\leftarrow$ tree.root.children
          \Comment{The set of ICD-10 chapters}
          \\
          \While{true}
            \State code\_descriptions $\leftarrow$ tree.get\_descriptions(candidate\_codes)
            \State llm\_response $\leftarrow$ llm\_api\_request(case\_note, code\_descriptions)
            \State predicted\_codes $\leftarrow$ match\_code\_descriptions(llm\_response, code\_descriptions)
            \\
            \For{code in predicted\_codes}
                \If{code in tree.assignable\_codes}
                    \State assigned\_codes.append(code)
                \Else
                    \State parent\_codes.append(code)
                \EndIf
            \EndFor
            \\
            \If{parent\_codes.length > 0}
                \State parent\_code $\leftarrow$ parent\_codes.pop(0)
                \State candidate\_codes $\leftarrow$ tree.get\_child\_codes(parent\_code)
            \Else
                \State break
                \Comment{There are no further codes to explore.}
            \EndIf
          \EndWhile
        
          \State \textbf{return} assigned\_codes
        \EndFunction
    \end{algorithmic}
\end{algorithm}

Under the simplifying assumption that generating a response to each prompt takes constant time, the time complexity for the LLM-guided tree-search algorithm is the same as that of a multi-label decision tree query: $\mathcal{O}(k \cdot \log(d))$, where $k$ refers to the number of predicted labels and and $d$ refers to the depth of the tree at which these labels are found.

Whilst any single path from the root of the tree to a leaf is on average $\approx4$ steps, if many paths are explored, the number of prompts can be large ($>100$). We therefore enforce a simple prompt limit (not described in Algorithm \ref{app:tree-search-algorithm}) in place of the \texttt{while true} loop to avoid excessive steps. We set this hyper-parameter to a value of $50$, as early experiments suggested that additional steps beyond this yield little benefit.

\newpage

\end{document}